\documentclass[conference]{IEEEtran}
\IEEEoverridecommandlockouts

\usepackage{amsmath,amssymb,amsfonts}
\usepackage{graphicx}
\usepackage{booktabs}
\usepackage[hidelinks]{hyperref}
\usepackage{flushend}

\newcommand{\trail}{TRAIL}
\newcommand{\dsam}{Detective SAM}

\begin{document}

\title{
Frozen DINO Localizes Image Edits Without a Localizer}

\author{
\IEEEauthorblockN{Zane Kumar}
\IEEEauthorblockA{\textit{Department of Computing}\\
\textit{St Paul's School London}\\
London, United Kingdom\\
{kumarz@stpaulsschool.org.uk}}
\and
\IEEEauthorblockN{Vishal Jain}
\IEEEauthorblockA{\textit{Department of Computing}\\
\textit{Imperial College London}\\
London, United Kingdom\\
{v.jain24@imperial.ac.uk}}
\and
\IEEEauthorblockN{Bernhard Kainz}
\IEEEauthorblockA{\textit{Computing / AIBE}\\
\textit{Imperial College London / FAU}\\
London, UK / Erlangen, Germany\\
{b.kainz@imperial.ac.uk}}}

    \maketitle

\begin{abstract}
Localized image edits can alter the meaning of a photograph while leaving most of it authentic, so forensic analysis must identify where an edit occurred rather than only flag the image. Recent training-free detectors approach manipulation detection by perturbing an image and measuring the resulting change in a frozen vision encoder, but collapse that response to one image-level score. We show that the discarded patch-level response is itself a localization map. Training-free Localization of AI-image Edits from patch-token Drift (\trail{}) applies one global Haar perturbation and maps cosine drift between corresponding DINO patch tokens. To set a reference for current supervised localization capability, we compare with \dsam{}, a mask-supervised SAM2-based localizer. On 80 source-disjoint CocoGlide test images, \trail{} reaches .903 patch AUROC versus .912 for \dsam{}. Its fixed-threshold Dice is lower (.619 versus .709), while a per-image oracle threshold raises the same \trail{} maps to .790. We then test whether the cue depends on artifacts from a generator. With every \trail{} setting transferred unchanged, Poisson image interpolation reaches .855 AUROC versus .864 for \dsam{}, showing that the spatial response persists when no generator is used. We also find a consistent depth pattern across sixteen DINO-family encoders: the best localization block lies at normalized depth .80--.94 in every model tested. A separate scale study shows that larger models increase the fake-minus-aligned-real AUROC gap more consistently than raw localization AUROC. Finally, an ablation of how drift is generated shows that shared image context matters: Haar AUROC falls from .903 with a global perturbation to .857 for local-in-canvas perturbations and .735 for independently encoded crops.  A strong late-layer localization signal is already present in frozen DINO patch tokens; how clearly it is revealed depends on the perturbation used and whether global context is preserved. Code is available at \url{https://github.com/VishalJ99/trail-image-edit-localization}.
\end{abstract}


\section{Introduction}

Generative image editors can replace a small region of a photograph while leaving the rest unchanged. An image-level detector can flag that something was edited, but it cannot say which content should no longer be trusted. Edit localization asks for that missing spatial information.

Recent training-free detectors use perturbation sensitivity in frozen visual representations. RIGID~\cite{rigid} compares DINOv2 embeddings before and after Gaussian noise, MINDER~\cite{minder} studies how the useful perturbation changes with image domain, and WaRPAD~\cite{warpad} attenuates high-frequency Haar components inside independently encoded crops. These methods produce one score per image. Frozen DINOv3 patch tokens also carry global authenticity cues for cross-generator image-level detection~\cite{huangdinov3}. DRIFT~\cite{drift} produces patch-wise perturbation maps, but trains one-class projection heads for generated-image detection rather than mask-level localization of partial edits.

Forgery localization is usually trained directly from masks. TruFor~\cite{trufor} learns to combine forensic cues for dense localization, while \dsam{}~\cite{detectivesam} couples perturbation-driven features to a learned segmentation pipeline. Training-free localization has also been studied through diffusion reconstruction: Zhang et al.~\cite{aaai_iml} compare conditional and unconditional reconstructions at selected diffusion steps. That route uses a generative model at inference. We focus on a narrow question: how much localization is already present in the perturbation response of a frozen general-purpose vision encoder? We study the DINO family of vision encoders~\cite{dino,dinov2,dinov3}, since it is one of the most commonly used general-purpose vision backbones.

In contrast to prior training-free detection methods, which collapse perturbation responses into an image-level score, we retain their spatial structure and read out the response directly at the patch level. Given an image and one globally perturbed copy, a frozen DINO encoder produces two aligned grids of patch tokens. Cosine drift between corresponding tokens forms a coarse edit map. We call this Training-free Localization of AI-image Edits from patch-token Drift (\trail{}). The method has no learned localization head, decoder, or segmentation model. ``Training-free'' here means that no model parameter is fitted to manipulation labels or masks; the encoder is pretrained, while the block, spatial filter, and operating threshold are selected on development data.

We use \dsam{}~\cite{detectivesam}, a recent mask-supervised localizer, to set a reference for supervised localization performance. On CocoGlide, \trail{} reaches .903 patch AUROC against .912 for \dsam{}, with a paired 95\% interval that crosses zero. To test whether this response is specific to generative artifacts, we transfer the same configuration to Poisson image interpolation, where no generator is used; \trail{} reaches .855 AUROC against .864. TGIF2 gives a harder regime: the random-mask AUROC gap is unresolved, while semantic-mask splices give the supervised model a .092 advantage.

In an encoder study, we find that across sixteen DINO-family encoders, the strongest localization block lies at normalized depth .80--.94 in every model tested. Scaling the encoder changes a different quantity more consistently: the fake-minus-real increment rises with model size in three of four families even when raw localization has already saturated. A separate perturbation ablation asks how the drift map should be generated. For Haar, AUROC falls from .903 with a global perturbation to .857 with local-in-canvas perturbations and .735 with independently encoded crops. The experiments therefore separate three claims: frozen patch drift can localize edits, the useful signal is consistently late, and readout quality depends on both perturbation choice and preserved global context.

\begin{figure*}[t]
\centering
\includegraphics[width=0.98\textwidth]{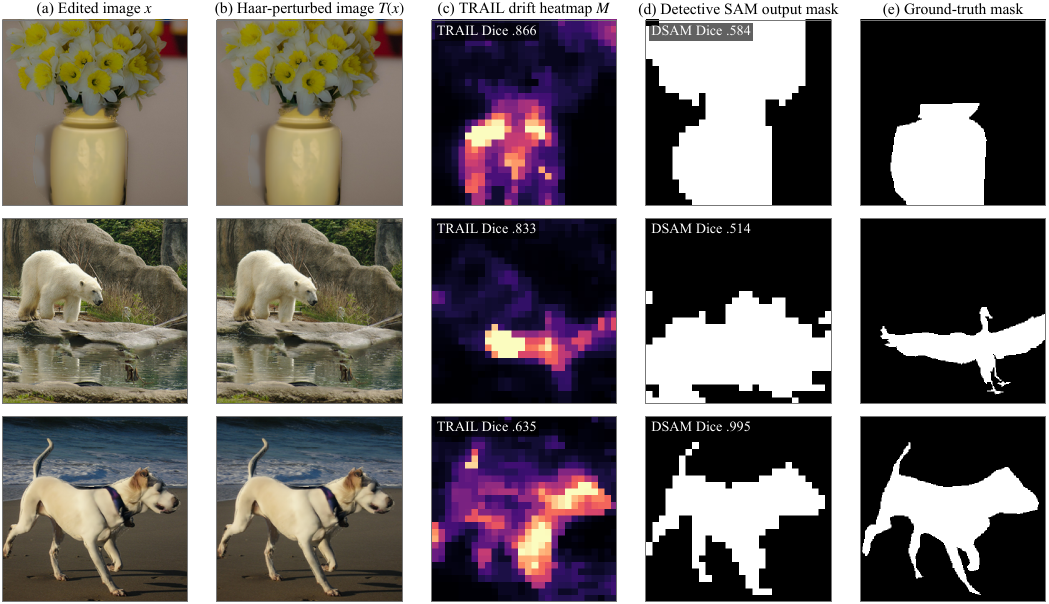}
\caption{Three illustrative CocoGlide examples; aggregate performance is
reported in Table~\ref{tab:main}. Columns show the edited input, the globally
Haar-attenuated input, the \trail{} patch-drift map, the binary \dsam{} output
at its released threshold, and the ground-truth mask. \trail{} maps are
normalized independently for display; brighter colors indicate greater
relative drift. Dice labels use each method's fixed operating point. Ground
truth is used only for evaluation.}
\label{fig:qual}
\end{figure*}

\section{Method}
\label{sec:method}

Let $f_{\ell,p}(x)\in\mathbb{R}^{d}$ denote patch token $p$ after transformer block $\ell$ of a frozen encoder. CLS and register tokens are excluded. No encoder parameter is updated. For a two-level Haar decomposition, let $\mathrm{HF}_2(x)$ denote the reconstructed detail component. We perturb the full image by
\begin{equation}
T(x)=x-\alpha\,\mathrm{HF}_2(x), \qquad \alpha=0.2.
\label{eq:perturb}
\end{equation}
The transform is deterministic and does not use the edit mask. Applying it to the full canvas keeps each token in the same surrounding image context in both encoder passes.

The score at patch $p$ is the cosine drift between the original and perturbed token,
\begin{equation}
s_{\ell}(p)=1-\frac{f_{\ell,p}(x)^\top f_{\ell,p}(T(x))}
{\lVert f_{\ell,p}(x)\rVert_2\lVert f_{\ell,p}(T(x))\rVert_2}.
\label{eq:drift}
\end{equation}
The scores are reshaped to the token grid and passed through one reflect-padded $3\times3$ median filter. The result is the \trail{} map $M$. There is no learned spatial post-processing.

The headline configuration uses DINOv3 ViT-7B/16~\cite{dinov3} at $448\times448$ and block 36, selected before test access by maximizing mean within-image patch AUROC over the 320 development sources. This gives a $28\times28$ map; binary masks use the development-selected threshold $\tau=0.0026$. We also report the diagnostic per-image oracle
\begin{equation}
\tau^\star(x)=\arg\max_t\mathrm{Dice}(\mathbb{1}[M\ge t],Y),
\label{eq:oracle}
\end{equation}
where $Y$ is the ground-truth mask. The oracle is unavailable at test time. It asks how much overlap can be recovered from the same spatial ordering if threshold calibration is solved separately for each image. Inference requires one wavelet transform and two frozen-encoder passes.

\section{Experimental setup}
\label{sec:setup}

\noindent\textbf{Data.} CocoGlide, released with TruFor~\cite{trufor}, contains GLIDE~\cite{glide} inpainting edits of COCO images with ground-truth masks. From 499 eligible sources, 400 were sampled by source into 320 development and 80 test images. The block, median filter, and operating threshold were fixed by source-level five-fold cross-validation before test scoring. TGIF2~\cite{tgif2} is used only as a transfer test. We evaluate 720 variation-0 splices from 80 authentic test sources: 400 semantic-mask edits across FLUX.1-dev, FLUX.1-fill-dev, FLUX.1-schnell, Photoshop Firefly, and SD2, and 320 random-mask edits across four generator conditions. Conditions are averaged within source before the final 80-source mean. Fully regenerated images are excluded because their masks cover semantic changes over most of the frame rather than a localized edit. We also transfer the frozen configuration to Foreign Patch Interpolation (FPI) and Poisson Image Interpolation (PII)~\cite{pii}, two classical edit families constructed without a generative model.

\noindent\textbf{Metrics and protocol.} Every method is scored on the same $28\times28$ grid. A patch is positive when edit-mask coverage is at least $0.5$, negative at zero coverage, and ignored at mixed boundaries. Patch AUROC measures spatial ranking. Dice at each method's prespecified operating point measures binary mask quality. For \trail{}, oracle Dice uses $\tau^\star$. Paired method gaps use 10,000 source-level bootstrap resamples. AUPRC is reported in the text where it distinguishes methods with similar AUROC.

\noindent\textbf{Reference and controls.} \dsam{} runs from its released checkpoint and released threshold of $0.5$; its output is resized and block-averaged to the same token grid. A raw-pixel control applies the same Haar perturbation and median filter to RMS pixel change. An aligned-real control runs \trail{} on the unedited source and applies the edit mask only after scoring. The first tests whether perturbation magnitude in pixel space explains the map. The second measures how much drift the same spatial region already has before editing.

\noindent\textbf{Encoder study.} We test sixteen encoders from DINO~\cite{dino}, DINOv2~\cite{dinov2}, DINOv2 with registers~\cite{registers}, and DINOv3, spanning ViT-S to 7B. Each model's block is selected on the development folds and evaluated on the same 80 CocoGlide test sources. For each encoder we report fake-image AUROC and the paired fake-minus-aligned-real AUROC increment. The increment subtracts the AUROC obtained when the same edit masks are placed over the authentic source. A larger value means that the spatial ranking is more specific to the edit rather than ordinary content in that region.

\begin{table*}[t]
\centering
\caption{Frozen \trail{} and mask-supervised \dsam{} are compared at matched
$28\times28$ resolution across generative and classical edits; higher values
are better.}
\label{tab:main}
\footnotesize
\setlength{\tabcolsep}{2.35pt}
\begin{tabular}{l r rrr rrr rrr r}
\toprule
& & \multicolumn{3}{c}{Patch AUROC}
& \multicolumn{3}{c}{Fixed-operating-point Dice}
& \multicolumn{3}{c}{Per-image oracle Dice} & \multicolumn{1}{c}{\trail{}} \\
\cmidrule(lr){3-5} \cmidrule(lr){6-8} \cmidrule(lr){9-11} \cmidrule(lr){12-12}
Dataset / regime & $N$ & \trail{} & Det. SAM & $\Delta$
& \trail{} & Det. SAM & $\Delta$ & \trail{} & Det. SAM & $\Delta$ & F--R \\
\midrule
CocoGlide & 80 & .903 & .912 & $-.009^{\dagger}$ & .619 & .709 & $-.090$ & .790 & --- & --- & $+.206$ \\
TGIF2 semantic & 400 & .811 & .904 & $-.092$ & .212 & .357 & $-.145$ & .363 & .540 & $-.178$ & $+.068$ \\
TGIF2 random & 320 & .727 & .758 & $-.031^{\dagger}$ & .256 & .281 & $-.026^{\dagger}$ & .405 & .408 & $-.003^{\dagger}$ & $+.164$ \\
FPI & 80 & .850 & .894 & $-.044^{\dagger}$ & .482 & .694 & $-.211$ & .697 & .829 & $-.131$ & $+.151$ \\
PII & 79 & .855 & .864 & $-.009^{\dagger}$ & .558 & .613 & $-.056$ & .769 & .800 & $-.032$ & $+.157$ \\
\bottomrule
\end{tabular}

\vspace{2pt}
\begin{minipage}{0.98\textwidth}
\scriptsize
\emph{Notes.} $N$ is the number of evaluable edited images; TGIF2 results are
macro-averaged over 80 authentic sources. $\Delta$ is \trail{} minus \dsam{};
$\dagger$ marks a paired 95\% source-bootstrap interval containing zero, while
the other displayed method gaps exclude zero. Fixed Dice uses each method's
prespecified operating point (\trail{} $\tau=.0026$; \dsam{} $\tau=.5$).
Oracle Dice selects a ground-truth-dependent threshold per image and is
diagnostic only; \dsam{} oracle Dice was not retained for CocoGlide. F--R is
\trail{}'s edited-minus-aligned-real localization-AUROC increment. FPI and PII
denote Foreign Patch Interpolation and Poisson Image Interpolation; both use
effective masks containing only pixels whose quantized RGB values change.
\end{minipage}
\end{table*}

\section{Results}
\label{sec:results}

\subsection{Frozen token drift}
\label{sec:main}

Table~\ref{tab:main} shows our main comparison. On CocoGlide, \trail{} reaches .903 AUROC and \dsam{} .912. The paired difference is $-.009$ with a 95\% interval of $[-.049,.033]$. AUPRC is .788 against .841, with a paired interval of $[-.113,.007]$. The raw drift map before median filtering reaches .843 AUROC, so the small spatial filter adds .060. The identically filtered raw-pixel control reaches .641 and the aligned-real control .697. The fake-minus-aligned-real increment for \trail{} is $+.206$ $[.167,.246]$. The localization signal therefore cannot be explained by stronger pixel-space perturbation inside the edited region or by ordinary content at the same location.

The clearer difference on CocoGlide is thresholded overlap. At their prespecified operating points, Dice is .619 for \trail{} and .709 for \dsam{}, a paired deficit of .090 $[.036,.146]$; \trail{} still has the higher per-image Dice on 35\% of the test set. Choosing the best threshold separately for each \trail{} map raises Dice to .790, an increase of .171 $[.134,.209]$. We did not measure a CocoGlide oracle for \dsam{}, so .790 is not an oracle comparison between methods. It measures calibration headroom in the frozen \trail{} maps. Fixed and oracle Dice remain correlated across images (Pearson .72, Spearman .62). The oracle thresholds span .00033 to .01342, with 44 of 80 images preferring a cut below the development value of .0026 and 36 preferring one above it. The development-frozen cut therefore misses in both directions. The soft map can remain well placed when one fixed cut over- or under-segments it, as in Fig.~\ref{fig:qual}.

Patch-grid resolution does not account for most of this gap. The best Dice available to any binary $28\times28$ patch map against each pixel mask averages .926 and exceeds .8 for 94\% of the test images. The four images that recover less than 35\% of their own grid ceiling even under the oracle all contain edits covering 2--8\% of the frame. There are 26 images in that area band, however, and the other 22 recover most of their grid ceiling. Small edit area is therefore associated with failure but does not determine it. Mask area correlates only moderately with Dice (Pearson .31, Spearman .45).

Table~\ref{tab:main} also tests edits made without a generator. With the DINOv3 block, Haar dose, filter, and CocoGlide threshold transferred unchanged, \trail{} reaches .855 AUROC on PII against .864 for \dsam{}, and .850 on FPI against .894. A generator is therefore not necessary for patch-token drift to carry spatial information. Fixed Dice is .558 against .613 on PII and .482 against .694 on FPI, showing that similar ranking need not imply equally shaped masks at a fixed operating point.

TGIF2 gives the main boundary condition. Across 720 splices, \trail{} reaches .774 AUROC, .316 AUPRC, and .231 Dice, versus .839, .415, and .324 for \dsam{}. Per-image oracle Dice remains lower, .382 against .481. The gap is concentrated on semantic masks: the AUROC deficit is .092 $[.047,.136]$ and the oracle-Dice deficit is .178 $[.117,.239]$. On random masks, the AUROC gap is .031 and its paired interval reaches zero; the paired AUPRC, fixed-Dice, and oracle-Dice intervals also include zero. The aligned-real control is .743 AUROC under semantic masks but .563 under random masks, and the fake-minus-real increment falls from $+.164$ on random masks to $+.068$ on semantic masks. Some of the response inside semantic object regions is therefore present before the edit. Because mask geometry, semantics, and generator also change between these groups, the experiment establishes a regime difference rather than a cause.

\begin{figure*}[t]
\centering
\includegraphics[width=0.98\textwidth]{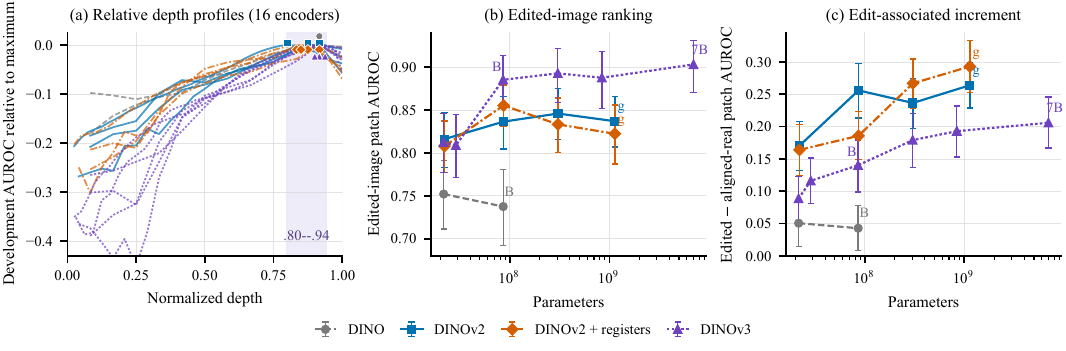}
\caption{Depth and scale across sixteen DINO-family encoders. (a) Development
AUROC relative to each model's own maximum; family-specific marker lanes show
all selected blocks, which lie at normalized depth .80--.94. (b) Edited-image
patch-AUROC at those blocks. (c) Paired edited-minus-aligned-real patch-AUROC,
not a real-versus-fake classifier. Panels (b,c) show 95\% source-bootstrap
intervals using identical source resamples for both metrics of each encoder. From DINOv3-B
to 7B, the point-estimate ranking difference is +.018 and the increment
difference is +.066.}
\label{fig:depth}
\end{figure*}

\subsection{Edit-specific scale vs. signal contrast}
\label{sec:depth}

Figure~\ref{fig:depth} shows that across all sixteen encoders, the development-selected block lies at normalized depth .80--.94. No model selects an early or middle block. This range holds across DINO, DINOv2, DINOv2 with registers, and DINOv3 despite their different sizes and pretraining recipes. The choice is also locally stable in the headline encoder: blocks 35, 36, and 37 of DINOv3 ViT-7B give .9029, .9037, and .9032 AUROC on development data. The late-layer result is therefore not created by a single narrow block optimum.

We find that model size changes the fake-minus-real increment more consistently than raw localization. In DINOv3, fake-image AUROC is .885, .893, .888, and .903 for ViT-B, L, H+, and 7B, so raw ranking changes little after ViT-B. The fake-minus-real increment rises through the family, from .089 at ViT-S to .116, .140, .179, .193, and .206 at 7B. DINOv2 with registers shows a similar separation: raw AUROC is .808, .855, .833, and .822 across S, B, L, and g, while the increment rises from .164 and .186 to .267 and .293 at the two larger models. Plain DINOv2 gives increments of .171, .256, .237, and .264 across S, B, L, and g. DINO is the exception: its two tested sizes have increments of .051 and .043, with raw AUROC .752 and .738.

At L and g, DINOv2 with registers gives .267 versus .237 and .293 versus .264 for plain DINOv2; at S and B the ordering reverses (.164 versus .171 and .186 versus .256). The larger-model result is consistent with register tokens absorbing high-norm artifact tokens~\cite{registers}, but the small-model reversal rules out a uniform register advantage. Raw ranking and edit-specific contrast also select different encoders. DINOv3 ViT-7B has the highest fake-image AUROC at .903, while DINOv2 ViT-g with registers has the largest fake-minus-real increment at .293. Capacity therefore does not simply raise the same localization score. In the families with a clear scaling trend, it more reliably separates edit-associated drift from drift already present in the authentic source.

\begin{table}[t]
\centering
\caption{The prespecified perturbation/application sweep is evaluated by patch
AUROC on CocoGlide and TGIF2.}
\label{tab:sweep}
\footnotesize
\setlength{\tabcolsep}{3.1pt}
\begin{tabular}{ll cc}
\toprule
Perturbation & Application & CocoGlide & TGIF2 dev. \\
\midrule
Haar attenuation & Global & \textbf{.903} & \textbf{.809} \\
 & Local canvas & .857 & .735 \\
 & Independent crops & .735 & .453 \\
\addlinespace[1.5pt]
Gaussian noise & Global & \textbf{.798} & \textbf{.481} \\
 & Local canvas & .738 & .418 \\
 & Independent crops & .527 & .226 \\
\addlinespace[1.5pt]
Gaussian blur & Global & \textbf{.830} & \textbf{.802} \\
 & Local canvas & .646 & .650 \\
 & Independent crops & .610 & .389 \\
\bottomrule
\end{tabular}

\vspace{2pt}
\begin{minipage}{0.96\columnwidth}
\scriptsize
\emph{Notes.} Bold marks the best application within each perturbation.
CocoGlide uses 80 test images; TGIF2 uses a source-disjoint development cohort
of 80 authentic sources and 400 semantic splices. Configuration ranks agree
across datasets (Spearman $\rho=.867$, $p=.0025$).
\end{minipage}
\end{table}

\subsection{Global context}
\label{sec:geometry}

Table~\ref{tab:sweep} changes the perturbation and where it is applied while holding the encoder, block, and filter fixed. For every perturbation, localization falls from the global condition to local-in-canvas regions and again to independently encoded crops. Haar moves from .903 to .857 to .735 AUROC. Noise moves from .798 to .738 to .527, and blur from .830 to .646 to .610. Haar is also the best perturbation within each geometry. The same ordering transfers to TGIF2: global, local-canvas, and independent-crop Haar reach .809, .735, and .453.

The three conditions separate perturbation locality from encoding context. The global condition perturbs and encodes the full image. The local condition perturbs only local regions but leaves them in their original full-image context. Independent crops remove the surrounding image during encoding. The largest loss occurs when that shared context is removed. An independently encoded patch can respond both to the edit and to the change in its surrounding context, which makes its drift less useful as a localization score.

This ordering differs from WaRPAD's image-level detection result. WaRPAD uses independently encoded patches and averages their scores~\cite{warpad}. A stitched version of its published DINOv2-L/14 construction, kept as a $4\times4$ map instead of averaged to one score, reaches .808 AUROC on the 45 CocoGlide images left evaluable by its coarse mask rule. In WaRPAD's Synthbuster ablation, the full patch-wise method reaches .834 AUROC while the Haar perturbation without its rescale-and-patch construction reaches .636~\cite{warpad}. The crop construction that helps image-level detection therefore does not give the best localization map here. For localization, corresponding tokens are best compared while they remain inside a common full-image context.

The exact Haar dose is less important than the application geometry. On development data, $\alpha\in\{.05,.1,.2,.4\}$ gives .900, .908, .904, and .889 AUROC. The selected $\alpha=.2$ lies inside this broad range rather than at an isolated optimum. The global-Haar configuration in Table~\ref{tab:sweep} is also the one transferred unchanged to TGIF2; no perturbation or geometry was selected on that benchmark.

\section{Limitations}

\trail{} produces a $28\times28$ patch map rather than full-resolution segmentation, and the headline configuration uses two passes through a 7-billion-parameter encoder. Smaller DINOv3 models approach its raw AUROC but have smaller fake-minus-real increments. The supervised reference is block-averaged to the same grid, so this comparison does not measure the value of its native high-resolution output. The per-image oracle consumes ground truth and is only a diagnostic; selecting a stable threshold across new image domains remains open. The method is training-free in parameters, not validation-free, because the block, filter, and threshold are selected on development masks. The evaluation covers inpainting-style local edits, TGIF2 splices, and two classical edit families. Global retouching, fully generated images, recompression, resizing, and adversarial post-processing are not tested. The TGIF2 semantic and random groups also change several factors at once, so their difference should not be read as a causal effect of semantics alone.

\section{Conclusion}

Image-level perturbation detectors discard a spatial response that can already localize edits. Reading cosine drift between aligned frozen DINO patch tokens gives .903 AUROC on CocoGlide against .912 for a supervised localizer, and the same configuration remains effective on generator-free PII. The signal is consistently late across sixteen DINO-family encoders; larger models increase edit-specific contrast more reliably than raw localization. Global perturbations preserve the map better than independently encoded crops. These results identify a localization signal inside the frozen encoder before a localization model is fitted.

\section*{Acknowledgment}

V. Jain is supported by the UKRI Centre for Doctoral Training AI4Health (EP/S023283/1). We acknowledge HPC resources from NHR@FAU (projects b143dc and b180dc), funded by federal and Bavarian state authorities; NHR@FAU hardware is partially funded by DFG project 440719683.

\end{document}